\documentclass[runningheads]{llncs}
\usepackage{makeidx}
\usepackage{graphicx}
\usepackage{amsmath,amssymb} % define this before the line numbering.

\usepackage{subfigure}
\usepackage[section]{algorithm}
\usepackage{algorithmic}
\usepackage{url}

\newcommand{\cD}{\mathcal{D}}

\newcommand{\cI}{\mathcal{I}}

\newcommand{\cL}{\mathcal{L}}

\newcommand{\vt}{\vartheta}

\newcommand{\st}{\text{subject to}\;}

\newcommand{\R}{\mathbb{R}}

\newcommand{\diag}{\mathrm{diag}}

\newcommand{\suml}[2]{\sum\nolimits_{#1}^{#2}}

\newcommand{\eps}{\varepsilon}

\newcommand{\scal}[2]{\left\langle #1,#2 \right\rangle}
\newcommand{\la}{\langle}
\newcommand{\ra}{\rangle}

\newcommand{\etal}{\emph{et al.}}
\newcommand{\eg}{\emph{e.g.}}
\newcommand{\ie}{\emph{i.e.}}
\newcommand{\vs}{\emph{vs.}}
\mainmatter

\def\DAGM12SubNumber{47}  
\titlerunning{Revisiting loss-specific training of filter-based MRFs for image restoration}
\authorrunning{Yunjin Chen, Thomas Pock, Ren{\'e} Ranftl and Horst Bischof}
\author{Yunjin Chen, Thomas Pock, Ren{\'e} Ranftl and Horst Bischof
\thanks{This work was supported by the Austrian Science Fund (project no. P22492) and 
the Austrian Research Promotion Agency (project no. 832366)}}
\institute{Institute for Computer Graphics and Vision, TU Graz}
\title{Revisiting loss-specific training of filter-based MRFs for image restoration}
\begin{document}

\maketitle

\begin{abstract}
It is now well known that Markov random fields (MRFs) are 
particularly effective for modeling image priors in low-level 
vision. Recent years have seen the emergence of two main approaches 
for learning the parameters in MRFs: (1) probabilistic learning 
using sampling-based algorithms and (2) loss-specific training based 
on MAP estimate. After investigating existing training approaches, 
it turns out that the performance of the loss-specific training has 
been significantly underestimated in existing work. In this paper, 
we revisit this approach and use techniques from bi-level 
optimization to solve it. We show that we can get a substantial gain 
in the final performance by solving the lower-level problem in the 
bi-level framework with high accuracy using our newly proposed algorithm. 
As a result, our trained model is on par with highly specialized image denoising 
algorithms and clearly outperforms probabilistically trained MRF 
models. Our findings suggest that for the loss-specific training scheme, 
solving the lower-level problem with higher accuracy is beneficial. 
Our trained model comes along with the additional 
advantage, that inference is extremely efficient. Our GPU-based 
implementation takes less than 1s to produce state-of-the-art performance.
\end{abstract}
%%%%%%%%% BODY TEXT
\vspace*{-1cm}
\section{Introduction and previous work}
\vspace*{-0.25cm}
Nowadays the MRF prior is quite popular for solving various inverse problems in image processing 
in that it is a powerful tool for modeling the statistics of natural images. Image models based on MRFs, 
especially higher-order MRFs, 
have been extensively studied and applied to image processing tasks such as image denoising~\cite
{RothFOE2009,GaoCVPR2010,TappenCVPR2009,GaoDAGM2012,TappenVariational,TappenGaussian}, 
deconvolution~\cite{SchmidtCVPR11}, inpainting~\cite{RothFOE2009,GaoCVPR2010,TappenCVPR2009}, 
super-resolution~\cite{ZhangZLH12}, etc.

Due to its effectiveness, higher-order filter-based MRF models using the framework of the Field of Experts (FoE)~\cite{RothFOE2009}, 
have gained the most attention. They are defined by a set of linear filters and the potential function. 
Based on the observation that responses of mean-zero linear filters 
typically exhibit heavy-tailed distributions~\cite{Huang1999_Statistics} on natural images, 
three types of potential functions have been investigated, including the 
Student-t distribution (ST), generalized Laplace distribution (GLP) and Gaussian scale mixtures (GSMs) function.

In recent years several training approaches have emerged to learn the parameters of the MRF models~\cite{HintonPOE2002,Weissgoodmodel2007,RothFOE2009,GaoCVPR2010,TappenCVPR2009,Barbu2009,DomkeAISTATS2012,GaoDAGM2012}.
Table~\ref{foelearningresults} gives a summary of several typical methods and the corresponding average 
denoising PSNR results 
based on 68 test images from Berkeley database with $\sigma = 25$ Gaussian noise.
\begin{table}[t!]
\vspace{-0.5cm}
\begin{center}
\begin{tabular}{|l|l|l|l|l|}
\hline
model & potential & training & inference & PSNR\\
\hline\hline
$5 \times 5$ FoE& ST\&Lap. & contrastive divergence & MAP, CG & 27.77\cite{RothFOE2009} \\
\cline{1-5}
$3 \times 3$ FoE& GSMs & contrastive divergence & Gibbs sampling & 27.95\cite{GaoCVPR2010} \\
\cline{1-5}
$5 \times 5$ FoE&  GSMs & persistent contrastive divergence & Gibbs sampling & 28.40\cite{GaoDAGM2012}\\
\cline{1-5}
$5 \times 5$ FoE&  ST & loss-specific(truncated optimization) & MAP, GD& 28.24\cite{Barbu2009}\\
\cline{1-5}
$5 \times 5$ FoE&  ST & loss-specific(truncated optimization) & MAP, lbfgs~\cite{lbfgs}& 28.39\cite
{DomkeAISTATS2012}\\
\cline{1-5}
$5 \times 5$ FoE&  ST & loss-specific(implicit differentiation)& MAP, CG & 27.86\cite{TappenCVPR2009}\\
\cline{1-5}
\hline
\end{tabular}
\end{center}\vspace{-0.15cm}
\caption{Summary of various typical MRF-based systems and the average denoising results on 68 test images~\cite{RothFOE2009}
 with $\sigma = 25$}\vspace{-1.cm}
\end{table}
\label{foelearningresults}
Existing training approaches typically fall into two main types: 
(1) probabilistic training using (persistent) contrastive divergence ((P)CD); 
(2) loss-specific training. 
Roth and Black~\cite{RothFOE2009} first introduced the concept of FoE and proposed an approach to learn the parameters of 
FoE model which uses 
a sampling strategy and the idea of CD to estimate the expectation value over the model distribution. 
Schmidt 
\etal~\cite{GaoCVPR2010} improved the performance of their previous FoE model~\cite{RothFOE2009} by changing 
(1) the potential function to GSMs and (2) the inference method from MAP estimate to Bayesian minimum mean squared error estimate (MMSE). The 
same authors present their latest results in~\cite{GaoDAGM2012}, where they achieve significant improvements by employing an improved learning 
scheme called PCD instead of previous CD. 

Samuel and Tappen~\cite{TappenCVPR2009} 
present a novel loss-specific training approach to learn 
MRF parameters under the framework of bi-level optimization~\cite{bileveloverview}. 
They use a plain gradient-descent technique to 
optimize the parameters, where the essence of this learning scheme - the gradients, are calculated by using implicit differentiation technique. 
Domke~\cite{DomkeAISTATS2012} and Barbu~\cite{Barbu2009} propose two similar approaches 
for the training of MRF model parameters also 
under the framework of bi-level optimization. Their methods 
are some variants of standard bi-level optimization method~\cite{TappenCVPR2009}. 
In the modified setting, the MRF model is trained in terms of results after optimization is truncated to a fixed number of iterations, \ie, they 
do not solve the energy minimization problem exactly; instead, they just run some specific optimization algorithm for a fixed number of steps. 

In a recent work~\cite{ECCV2012RTF}, the bi-level optimization technique is employed to train a non-parametric 
image restoration framework based on Regression Tree Fields (RTF), resulting a new state-of-the-art. This technique is also 
exploited for learning the so-called analysis sparsity priors~\cite{FadiliAnalysisLearning}, which is somewhat related to the FoE model. 
\vspace*{-0.5cm}
\section{Motivation and contributions}
\vspace*{-0.3cm}
\textbf{Arguments}: The loss-specific training criterion is formally expressed 
as the following bi-level optimization problem
\begin{equation}\label{learningmodel}
\begin{cases}
\arg\min\limits_{\vt}L(x^*(\vt),g)\\
\st x^*(\vt) = \arg\min\limits_{x}E(x,f,\vt). 
\end{cases}
\end{equation}

The goal of this model 
is to find the optimal parameters $\vt$ to minimize the loss function $L(x^*(\vt),g)$, which is called the upper-level 
problem in the bi-level framework. The MRF model is defined by the energy minimization 
problem $E(x,f,\vt)$, which is called the lower-level problem. 
The essential point for solving this bi-level optimization problem is to calculate the gradient 
of the loss function $L(x^*(\vt),g)$ with respect to the parameters $\vt$. 
As aforementioned, \cite{TappenCVPR2009} employs the implicit differentiation technique 
to calculate the gradients explicitly; in contrast, \cite{DomkeAISTATS2012} and \cite{Barbu2009} 
make use of an approximation approach based on truncated optimization. 
All of them use the same ST-distribution as potential function; however, the latter two 
approaches surprisingly obtain much better performance than the former, as can be seen in Table~\ref{foelearningresults}. 

In principle, Samuel and Tappen should achieve better (at least similar) results compared to 
the approximation approaches, because they use a ``full'' fitting training scheme, 
but actually they fail in practice. Therefore, we argue 
that there must exist something imperfect in their training scheme, and we believe that 
we will very likely achieve noticeable improvements by refining this ``full'' fitting training scheme. 

\textbf{Contributions}: Motivated by the above investigation, 
we think it is necessary and worthwhile to restudy the loss-specific training scheme and we expect that 
we can achieve significant improvements. 
In this paper, we do not make any modifications to the training model used in~\cite{TappenCVPR2009} - 
we use exactly the same model capacity, potential function and training images. The only difference is 
the training algorithm. We exploit a refined training algorithm that we solve the lower-level problem in the 
loss-specific training with very high accuracy and 
make use of a more efficient quasi-Newton's method for model parameters optimization. 
We conduct a series of playback experiments and we show that the performance of loss-specific training 
is indeed underestimated in previous work~\cite{TappenCVPR2009}. 
We argue that the the critical reason is that they have not solved the lower-level problem 
to sufficient accuracy. We also demonstrate that 
solving the lower-level problem with higher accuracy is indeed beneficial. 
This argument about the loss-specific training scheme is the major contribution of our paper. 

We further show that our trained model can obtain slight improvement by increasing the model size. 
It turns out that for image denoising task, our optimized MRF (opt-MRF) model of size $7 \times 7$ has 
achieved the best result among existing MRF-based systems and been on par 
with state-of-the-art methods. 
Due to the simplicity of our model, it is easy to implement the inference algorithm on parallel computation 
units, \eg, GPU. Numerical results show that our GPU-based implementation can perform image denoising 
in near real-time with clear state-of-the-art performance. 

\vspace*{-0.25cm}
\section{Loss-specific training scheme: bi-level optimization}
\vspace*{-0.25cm}
In this section, we firstly present the loss-specific training model. 
Then we consider the optimization problem from a more general point of view. Our derivation 
shows that the implicit differentiation technique employed in previous work~\cite{TappenCVPR2009} 
is a special case of our general formulation.  
\vspace*{-0.25cm}
\subsection{The basic training model}
\vspace*{-0.1cm}
Our training model makes use of the bi-level optimization framework, and is conducted based on the image denoising task. 
For image denoising, the ST-distribution based MRF model is expressed as
\begin{equation}\label{MRFmodel} 
\arg\min\limits_{x}E(x) = 
\suml{i=1}{N_f}{\alpha_i}\suml{p=1}{N_p}\rho((K_ix)_p) + \frac{\lambda}{2}\|x- f \|_2^2. 
\end{equation}

This is the lower-level problem in the bi-level framework. 
Wherein $N_f$ is the number of filters, $N_p$ is the number of pixels in image $x$, 
$K_i$ is an $N_p \times N_p$ 
highly sparse matrix, which makes the convolution of the filter $k_i$ with a two-dimensional image $x$ equivalent to
the product result of the matrix $K_i$ with the vectorization form of $x$, \ie, $k_i * x \Leftrightarrow K_i x$. 
In our training model, we express the filter $K_i$ as a linear combination of a set of basis filters $\{B_1,\cdots,B_{N_B}\}$, \ie, 
$K_i = \suml{j=1}{N_B}\beta_{ij}B_j$. 
Besides, $\alpha_i \geq 0$ is the parameters of ST-distribution for filter $K_i$, and 
$\lambda$ defines the trade-off between the prior term and data fitting term. 
$\rho(\cdot)$ denotes the Lorentzian potential function $\rho(z) = \text{log}(1 + z^2)$, 
which is derived from ST-distribution. 

The loss function $L(x^*,g)$ (upper-level problem) is defined to measure the difference between the optimal solution of 
energy function and the 
ground-truth. In this paper, we make use of the same loss function as in~\cite{TappenCVPR2009}, 
$L(x^*,g) = \frac{1}{2}\|x^* - g\|_2^2$, 
where $g$ is the ground-truth image and $x^*$ is the minimizer of \eqref{MRFmodel}. 

Given the training samples $\{f_k,g_k\}_{k=1}^N$, where $g_k$ and $f_k$
are the $k^{th}$ clean image and the associated noisy version respectively, our aim is to learn an optimal MRF parameter 
$\vt = (\alpha, \beta)$ (we group the coefficients $\beta_{ij}$ and weights $\alpha_i$ into a single vector $\vt$), 
to minimize the overall loss function. Therefore, the learning model is formally formulated as the
following bi-level optimization problem
\begin{equation}\label{learningmodel}
\begin{cases}
\min\limits_{\alpha \geq 0, \beta}L(x^*(\alpha,\beta)) = 
\suml{k=1}{N}\frac{1}{2}\|x_k^*(\alpha,\beta) - g_k\|_2^2\\
\text{where}~
x_k^*(\alpha,\beta) = \arg\min\limits_{x}
\suml{i=1}{N_f}\alpha_i\rho(K_ix) + \frac{1}{2}\|x- f_k \|_2^2, 
\end{cases}
\end{equation}
where $\rho(K_ix) = \suml{p=1}{N_p}\rho((K_ix)_p)$. 
We eliminate $\lambda$ for simplicity, since it can be incorporated into weights $\alpha$. 
%-------------------------------------------------------------------------
\vspace*{-0.25cm}
\subsection{Solving the bi-level problem}
\vspace*{-0.1cm}
In this paper, we consider the bi-level optimization problem from a general point of view. 
In the following derivation we only consider the case of a single 
training sample for convenience, and we show how to extend the framework to multiple training samples in the end. 

According to the optimality condition, the solution of the lower-level problem in~\eqref{learningmodel} is 
given by $x^*$, such that $\nabla_xE(x^*) = 0$. Therefore, we can rewrite problem~\eqref{learningmodel} as 
following constrained optimization problem
\begin{equation}\label{constrainedmodel}
\begin{cases}
\min\limits_{\alpha \geq 0, \beta}L(x(\alpha, \beta)) = 
\frac{1}{2}\|x(\alpha, \beta) - g\|_2^2\\
\st
\nabla_xE(x) = \suml{i=1}{N_f}\alpha_i K_i^T\rho'(K_ix) + x- f = 0, 
\end{cases}
\end{equation}
where $\rho'(K_ix) = (\rho'((K_ix)_1),\cdots,\rho'((K_ix)_p))^T \in \R^{N_p}$. 
Now we can introduce Lagrange multipliers and study the Lagrange function 
\vspace*{-0.25cm}
\begin{equation}\label{Lagrange}
\cL(x,\alpha,\beta,p,\mu) = 
\frac{1}{2}\|x - g\|_2^2 + \scal{-\alpha}{\mu} + \la \suml{i=1}{N_f}\alpha_i K_i^T\rho'(K_ix) + x- f,p\ra, 
\end{equation}
where $\mu \in \R^{N_f}$ and $p \in \R^{N_p}$ are the Lagrange multipliers associated to 
the inequality constraint $\alpha \geq 0$ and the equality constraint 
in~\eqref{constrainedmodel}, respectively. 
Here $\scal{\cdot}{\cdot}$ denotes the standard inner product. 
Taking into account the inequality constraint $\alpha \geq 0$, the first order necessary condition for optimality is given by 
\begin{equation}\label{optimality}
G(x,\alpha,\beta,p,\mu) = 0,
\end{equation}
where
\begin{equation*}
G(x,\alpha,\beta,p,\mu)=
\begin{pmatrix}
    (\suml{i=1}{N_f}\alpha_i K_i^T\cD_i K_i+ \cI)p + x-g\\[1.7ex]
    (\la K_i^T\rho'(K_ix),p \ra)_{N_f \times 1}-\mu\\[1.7ex]
    (\la B_j^T\rho'(K_ix) + K_i^T\cD_i B_jx, p \ra)_{n \times 1}\\[1.7ex]
     \suml{i=1}{N_f}\alpha_i K_i^T\rho'(K_ix) + x- f\\[1.7ex]
    \mu - \max(0,\mu-c\alpha)
\end{pmatrix}.
\end{equation*}
Wherein $\cD_i(K_ix) = \diag (\rho''((K_ix)_1),\cdots,\rho''((K_ix)_p)) \in \R^{N_p \times N_p}$, 
$(\la \cdot ,p \ra)_{N \times 1} = (\la (\cdot)_1,p \ra,\cdots,\la (\cdot)_r,p \ra)^T$, in the third formulation $n = N_f \times N_B$. 
Note that the last formulation is derived from the optimality condition for the inequality constraint $\alpha \geq 0$, which is expressed 
as $\alpha \geq 0, \mu \geq 0, \la \alpha, \mu \ra = 0$. It is easy to check that these three conditions are equivalent to 
$\mu - \max(0,\mu-c\alpha) = 0$ with $c$ to be any positive scalar and max operates coordinate-wise. 

Generally, we can continue to calculate the generalized Jacobian of G, \ie, the Hessian matrix of Lagrange function, 
with which we can then employ a Newton's method to solve the necessary optimality system~\eqref{optimality}. 
However, for this 
problem calculating the Jacobian of G is computationally intensive; thus in this paper we do not consider it and only make use of 
the first derivatives. 

Since what we are interested in is the MRF parameters $\vt=\{\alpha,\beta\}$, we can reduce unnecessary variables in \eqref{optimality}. By solving for 
$p$ and $x$ in~\eqref{optimality}, and substituting them into the second and the third formulation, we arrive at the gradients of 
loss function with respect to parameters $\vt$ 
\vspace*{-0.25cm}
\begin{equation}\label{overallderivative}
\begin{cases}
\nabla_{\beta_{ij}} L = 
- (B_j^T\rho'(K_ix) + K_i^T\cD_i B_jx)^T(H_E(x))^{-1}
\left(x - g\right)
\\
\nabla_{\alpha_{i}} L = 
- (K_i^T\rho'(K_ix))^T(H_E(x))^{-1}
\left(x - g\right)
\\
\text{where} ~\nabla_xE(x) = \suml{i=1}{N_f}\alpha_i K_i^T\rho'(K_ix) + x- f = 0. 
\end{cases}
\end{equation}
In~\eqref{overallderivative}, $H_E(x)$ denotes the Hessian matrix of $E(x)$, 
\begin{equation}\label{hessian}
H_E(x) = \suml{i=1}{N_f}\alpha_i K_i^T\cD_i K_i+ \cI.
\end{equation}

In~\eqref{overallderivative}, we also eliminate the Lagrange multiplier $\mu$ associated to 
the inequality constraint $\alpha \geq 0$, as we utilize a quasi-Newton's method for optimization, 
which can easily handle this type of box constraints. 
We can see that \eqref{overallderivative} is 
equivalent to the results presented in previous work~\cite{TappenCVPR2009} using implicit differentiation. 

Considering the case of $N$ training samples, in fact it turns out 
that the derivatives of the overall loss function in \eqref{learningmodel} with respect to the parameters $\vt$ are just the sum of 
\eqref{overallderivative} over the training dataset.

As given by \eqref{overallderivative}, we have collected all the necessary information to compute the required gradients, 
so we can now employ gradient descent based algorithms for optimization, \eg, steepest-descent algorithm. 
In this paper, we turn to 
a more efficient non-linear optimization method--the LBFGS quasi-Newton's method~\cite{lbfgs}. 
In our experiments, we will make use of the LBFGS implementation distributed 
by L. Stewart\footnote{\url{http://www.cs.toronto.edu/~liam/software.shtml}}.
In our work, the third equation in \eqref{overallderivative} is completed the L-BFGS algorithm, 
since this problem is smooth, to which 
L-BFGS is perfectly applicable. The training algorithm is terminated when the relative change of the loss 
is less than a tolerance, \eg, $tol = 10^{-5}$ or a maximum number of iterations \eg, $maxiter = 500$ is reached 
or L-BFGS can not find a feasible step to decrease the loss.
%------------------------------------------------------------------------
\vspace*{-0.5cm}
\section{Training experiments}
\vspace*{-0.3cm}
In order to demonstrate that the loss-specific training scheme was undervalued 
in previous work~\cite{TappenCVPR2009}, we conducted a playback experiment using (1) the same 40 images for 
training and 68 images for testing; (2) the same model capacity--24 filters of size $5 \times 5$; (3) the same basis
--``inverse'' whitened PCA \cite{RothFOE2009}, as in Samuel and Tappen's experiments. We randomly sampled four $51 \times 51$ patches 
from each training image, resulting in a total of 160 training samples. 
We then generated the noisy versions by adding Gaussian noise with standard deviation $\sigma = 25$. 

The major difference between our training experiment and previous one is the training algorithm. 
In our refined training scheme, we employed (1) our proposed algorithm to solve the lower-level problem with very high 
accuracy, and (2) LBFGS to optimize the model parameters, but in contrast, Samuel and Tappen used 
non-linear conjugate gradient and plain gradient descent algorithm, respectively. In our refined training algorithm, 
we used the normalized norm of the gradient, \ie, $\frac{\|\nabla_xE(x^*)\|_2}{\sqrt{N}} \leq \eps_l$ 
($N$ is the pixel number of the training patch) as the stopping criterion for solving the lower-level problem. 
In our training experiment, we set $\eps_l = 10^{-5}$ (gray-value in range [0 255]), which implies a very 
accurate solution. 

Based on this training configuration, we learned 24 filters of size $5 \times 5$, then we applied them 
to image denoising task to estimate the inference performance using the same 68 test images. Finally, we got 
an average PSNR value of 28.51dB for noise level $\sigma = 25$, which is significantly superior to previous 
result of 27.86dB in~\cite{TappenCVPR2009}. We argue that the major reason lies in 
our refined training algorithm that we solve the lower-level problem with very high accuracy. 

To make this argument more clear, we need to eliminate the possibility of 
training dataset, because we did not exploit exactly the same training dataset as previous work 
(unfortunately we do not have their dataset in hand). 
Since the training patches were randomly selected, we could run the training experiment multiple times by using 
different training dataset. Finally, we found that the deviation of test PSNR values based on 68 test images is within 0.02dB, 
which is negligible. Therefore, it is clear that training dataset is not the reason for this improvement, and the only 
remaining reason is our refined training scheme. 

\begin{figure}[t!]
\begin{center}
    \includegraphics[width=0.3\textwidth]{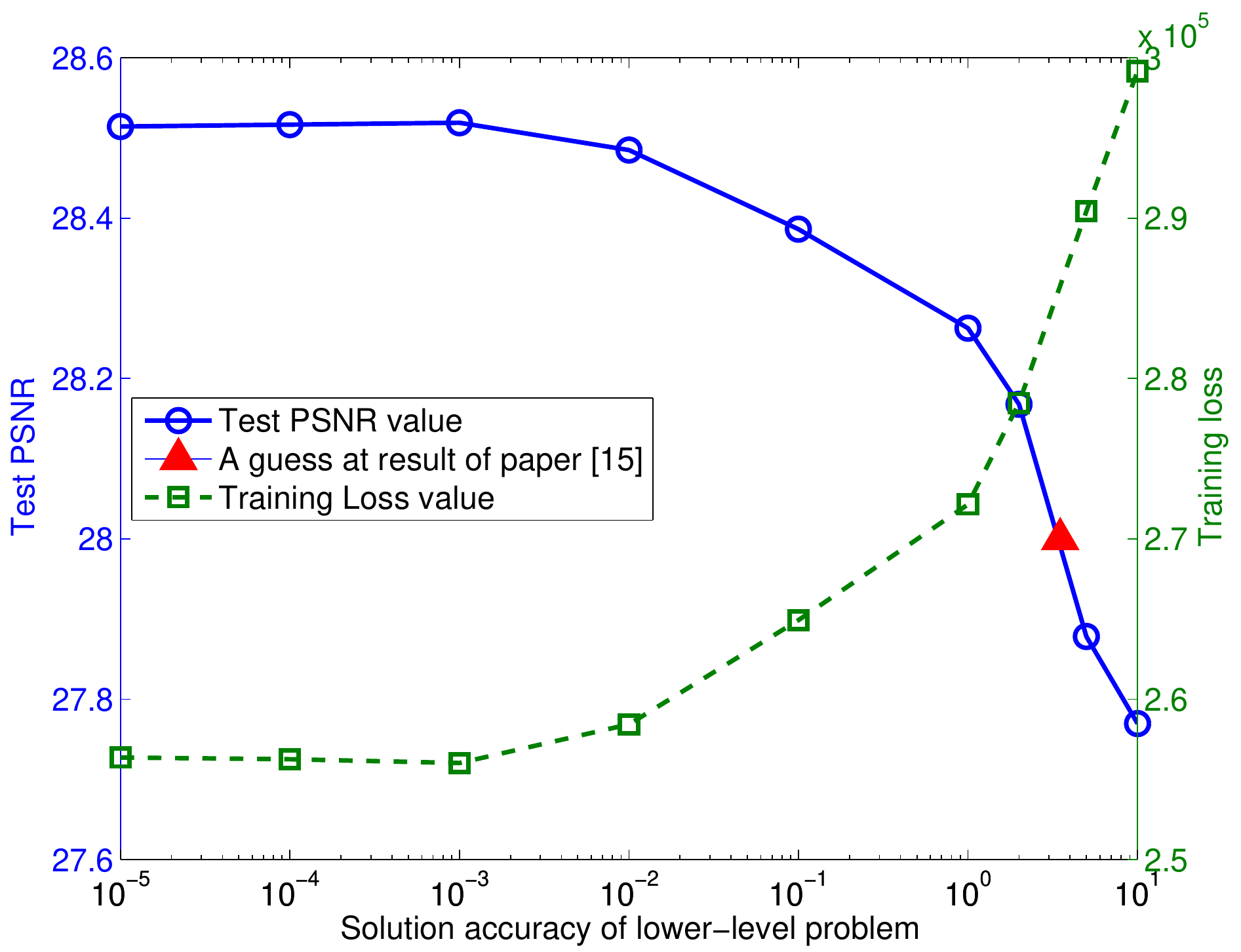}
    \includegraphics[width=0.3\textwidth]{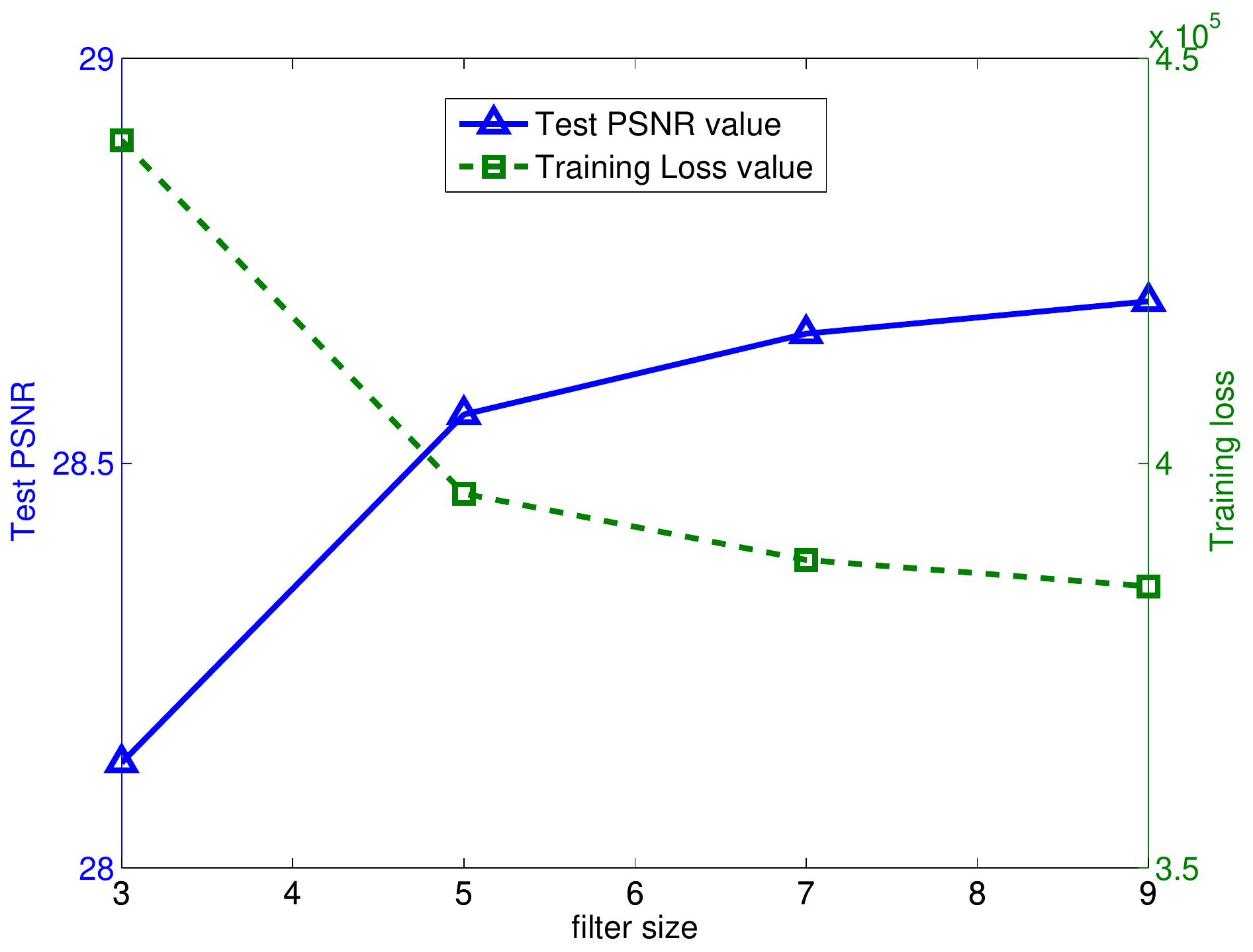}
\end{center}\vspace{-0.5cm}
    \caption{Performance curves (test PSNR value and training loss value) \vs 
\{the solution accuracy of the lower-level problem $\eps_l$ \& the filter size\}. 
It is clear that solving the lower-level problem with higher accuracy is beneficial and larger filter size can 
normally bring some improvement.}\vspace{-0.75cm}
\label{accuracycurve}
\end{figure}

%-------------------------------------------------------------------------------------------------
\textbf{The influence of $\eps_l$:} To investigate the influence of the solution accuracy of the lower-level 
problem $\eps_l$ more detailedly, 
we conducted a series of training and testing experiments by setting $\eps_l$ to different magnitudes. 
Based on a fixed training dataset (160 patches of size $51 \times 51$) and 68 test images, we got the performance 
curves with respect to the solution accuracy $\eps_l$, as shown in Figure~\ref{accuracycurve} (left). 
{From Figure~\ref{accuracycurve} (left), we can clearly see that it is indeed the high solution accuracy that helps us to 
achieve the above siginificant improvement. This finding is the main contribution of our paper.} 
We also make a guess how accurate Samuel and Tappen solve the lower-level 
problem according to their result and our performance curve, which is marked 
by a red triangle in Figure~\ref{accuracycurve} (left). 
The argument that higher solution accuracy of the lower-level problem is helpful is explicable, the reason is described below. 

As we know, the key aspect of our approach is to calculate the gradients of the loss function with respect to the parameters 
$\vt$. According to~\eqref{overallderivative}, there is a precondition to obtain accurate gradients: both the lower-level problem 
and the inverse matrix of Hessian matrix $H_E$ must be solved with high accuracy, \ie, we need to calculate a $x^*$ 
such that $\nabla_xE(x^*) = 0$ and compute $(H_E)^{-1}$ explicitly. 
Since the Hessian matrix $H_E$ is highly sparse, we can solve the linear system $H_E x = b$ efficiently with very high 
accuracy (we use the ``backslash'' operator in Matlab). 
However, for the lower-level problem, in practice we can only solve it to finite 
accuracy by using certain algorithms, \ie, $\frac{\|\nabla_xE(x^*)\|_2}{\sqrt{N}} \leq \eps_l$. 
If the lower-level problem is not solved to sufficient accuracy, the gradients $\nabla_\vt L$ are certainly inaccurate 
which will probably affect the training performance. This has been demonstrated in our experiments. 
Therefore, for the bi-level training framework, it is necessary to solve the lower-level problem as 
accurately as possible, \eg, in our training we solved it to a very high accuracy with $\eps_l = 10^{-5}$. 

\textbf{The influence of basis:} In our playback experiments, we used the ``inverse'' whitened PCA basis to keep 
consistent with previous work. However, we argue that the DCT basis is a better choice, because 
meaningful filters should be mean-zero according to the findings in~\cite{Huang1999_Statistics}, 
which is guaranteed by DCT basis without the constant basis vector. Therefore, we will 
exploit the DCT filters excluding the filter with uniform entries from now on. Using this modified DCT basis, we retrained our model and 
we got a test PSNR result of 28.54dB. 

\textbf{The influence of training dataset:} To verify whether larger training dataset is beneficial, we retrained 
our model by using (1) 200 samples of size $64 \times 64$ and (2) 200 samples of size $100 \times 100$, which is 
about two times and four times larger than our previous dataset, respectively. Finally, we got a test PSNR result 
of 28.56dB for both cases. As shown before, the influence of training dataset is marginal. 

\begin{figure}[t!]
%\vspace{-0.5cm}
  \begin{center}
    \includegraphics[width=0.98\textwidth]{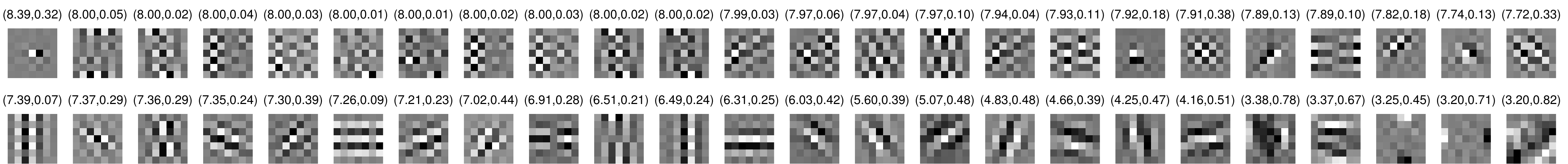}
\vspace{-0.25cm}
    \caption{48 learned filters ($7 \times 7$). 
The first number in the bracket is the weight $\alpha_i$ and the second one is the norm of the filter.}\vspace{-0.85cm}
\label{fig:DCT7}
  \end{center}
\end{figure}

\textbf{The influence of model capacity:} In above experiments, we concentrated on the model of size $5 \times 5$ to 
keep consistent with previous work. We can also train models of different filter sizes, \eg, $3 \times 3$, 
$7 \times 7$ or $9 \times 9$, to investigate the influence of model capacity. Based on 
the training dataset of 200 patches of size $64 \times 64$, we retrained our model with different filter size; the training results 
and testing performance are summarized in Figure~\ref{accuracycurve} (right). We can see that 
normally increasing the filter size can bring some improvement. However, the improvement of filter size $9 \times 9$ is 
marginal compared to filter size $7 \times 7$, yet the former is much more time consuming. The training time for the model with 
48 filters of size $7 \times 7$ was approximately 24 hours on a server (Intel X5675, 3.07GHz, 24 cores), but in contrast, the model 
of size $9 \times 9$ took about 20 days. More importantly, the inference time of the model of size $9 \times 9$ is 
certainly longer than the model of size $7 \times 7$, in that it involves more filters of larger size. Therefore, 
the model of size $7 \times 7$ offers the best trade-off between speed and quality, and we use it for 
the following applications. 
The learned 48 filters together with their associated weights and norms 
are presented in Figure~\ref{fig:DCT7}. 
\vspace{-0.5cm}
\section{Application results}
\vspace{-0.25cm}
An important question for a learned prior model is how well it generalizes. 
To evaluate this, we directly applied the above 48 filters of size $7 \times 7$ trained based on image denoising task to 
various image restoration problems such as image deconvolution, inpainting and super-resolution, as well as denoising. 
Due to space limitation, here we only present denoising results and the comparison to state-of-the-arts. 
The other results will be shown in the final version~\cite{gpu4vision}. 

%---------------------------------------------------------------------------
We applied our opt-MRF model to image denoising problem and compared its performance with leading image 
denoising methods, including three state-of-the-art methods: (1) BM3D~\cite{BM3D}; (2) LSSC~\cite{NLSC}; 
(3) GMM-EPLL~\cite{EPLL} along with two leading generic methods: (4) a MRF-based approach, FoE~\cite{GaoDAGM2012}; 
and (5) a synthesis sparse representation based method, KSVD~\cite{KSVDdenoising2006} trained on natural image patches. 
All implementations were downloaded from the corresponding authors' homepages. 
We conducted denoising experiments over 68 test images with various noise levels $\sigma = \{15, 25, 50\}$. 
To make a fair comparison, we used exactly the same noisy version of each test 
image for different methods and different test images were added with distinct noise realizations. 
All results were computed per image and then averaged over the test dataset. 
We used L-BFGS to solve the MAP-based MRF model~\eqref{MRFmodel}. 
When \eqref{MRFmodel} is applied to various noise level $\sigma$, 
we need to tune the parameter $\lambda$ (empirical choice $\lambda = 25/\sigma$).

\begin{table}[t!]
\begin{center}
\begin{tabular}{|l|c|c|c|c|c|c|}
\hline
$\sigma$ & KSVD & FoE & BM3D & LSSC & EPLL & Ours\\
\hline\hline
15 & 30.87 & 30.99 & 31.08 & \textbf{31.27} & 31.19 & 31.18\\
\cline{1-7}
25 & 28.28 & 28.40 & 28.56 & \textbf{28.70} & \textbf{28.68} & \textbf{28.66}\\
\cline{1-7}
50 & 25.17 & 25.35 & 25.62 & \textbf{25.72} & \textbf{25.67} & \textbf{25.70}\\
\cline{1-7}
\end{tabular}
\end{center}\vspace{-0.15cm}
\caption{Summary of denoising experiments results (average PSNRs over 68 test images from the Berkeley database). 
We highlighted the state-of-the-art results.}\vspace{-0.5cm}
\label{comparison}
\end{table}

Table~\ref{comparison} shows the summary of results. It is clear that our opt-MRF model 
outperforms two leading generic methods and has been on par with three state-of-the-art methods for any noise level. 
Comparing the result of our opt-MRF model 
with results presented in Table~\ref{foelearningresults}, our model has obviously achieved the best performance among 
all the MRF-based systems. 
To the best of our knowledge, this is the first time that a MRF model based on generic priors of natural images has 
achieved such clear state-of-the-art performance. We provide image denoising examples 
in the final version~\cite{gpu4vision}. 

In additional, our opt-MRF model is well-suited to GPU parallel computation in that it 
only contains the operation of convolution. Our GPU implementation based on 
NVIDIA Geforce GTX 680 accelerates the inference procedure significantly; for a denoising task with $\sigma = 25$, typically it takes 
0.42s for image size $512 \times 512$, 0.30s for $481 \times 321$ and 0.15s for $256 \times 256$. 
In Table~\ref{runningtime}, we show the average run time of the considered 
denoising methods on $481 \times 321$ images. 
Considering the speed and quality of our model, it is a perfect choice of the base methods in the image restoration framework 
recently proposed in~\cite{ECCV2012RTF}, which leverages advantages of existing methods. 

\begin{table}[t!]
\begin{center}
\hspace*{-0.25cm}
\begin{tabular}{r|c|c|c|c|c|c}
\cline{1-7}
 & KSVD & FoE & BM3D & LSSC & EPLL & Ours\\
\hline\hline
T(s) & 30 & 1600%\protect \footnotemark 
& 4.3 & 700 & 99 & 12 (\textbf{0.3})\\

psnr & 28.28 & 28.40 & 28.56 & {28.70} & {28.68} & {28.66}\\
\cline{1-7}
\end{tabular}
\end{center}\vspace{-0.15cm}
\caption{Typical run time of the denoising methods for a $481 \times 321$ image ($\sigma = 25$) on a server 
(Intel X5675, 3.07GHz). The highlighted number is the run time of GPU implementation.}\vspace{-1cm}
\label{runningtime}
\end{table}

%\footnotetext{This result was inferred based on the runtime of FoE model in~\cite{GaoCVPR2010} 
%according to the runtime statement in~\cite{GaoDAGM2012}.}
\vspace*{-0.5cm}
\section{Conclusion}
\vspace*{-0.25cm}
In this paper, we revisited the loss-specific training approach proposed by Samuel and Tappen 
in~\cite{TappenCVPR2009} by using a refined training algorithm. 
We have shown that the performance of the loss-specific training was indeed undervalued in previous work. 
We argued that the major reason lies in the solution accuracy of the lower-level problem 
in the bi-level framework, and we have demonstrated that solving the lower-level problem
with higher accuracy is beneficial. We have shown that we can further improve 
the performance of the learned model a little bit by using larger filters. 
For image denoising task, our learned opt-MRF model of size $7 \times 7$ 
presented the best performance among existing MRF-based systems, 
and has already been on par with state-of-the-art denoising methods. 
The performance of our opt-MRF model proves two issues: (1) the loss-specific training scheme under the 
framework of bi-level optimization, which is convergence guaranteed, is highly effective for parameters learning; 
(2) MAP estimate should be still considered as one of the leading approaches in low-level vision.

\vspace*{-0.5cm}
{\tiny
\bibliographystyle{splncs03}
\bibliography{iccv_bib}
}

\end{document}